\documentclass{article}
\PassOptionsToPackage{prologue,dvipsnames}{xcolor}

\usepackage{arxiv}
\usepackage{mathpazo}
\usepackage{hyperref}
\usepackage{url}

\usepackage[round]{natbib}
\usepackage{xspace}
\usepackage{amsmath, bm}
\usepackage{listings}
\usepackage{mathtools}
\usepackage{siunitx}
\usepackage{xspace}
\usepackage{ntheorem}
\usepackage{graphicx}
\usepackage{booktabs}
\usepackage{bbm}
\usepackage{stmaryrd}
\usepackage{nth}
\usepackage{subcaption}
\usepackage{tikz}
\usepackage{pgfplots}
\usepackage{setspace}

\usepackage{multirow}
\usepackage{placeins}
\usepackage{graphicx}
\usepackage{soul}
\usepackage[linesnumbered,ruled,vlined]{algorithm2e}
\usepackage{algpseudocode}
\SetKwInput{KwInput}{Input}
\usepackage{lipsum}
\usepackage{seqsplit}
\usepackage{multicol}
\usepackage{adjustbox}
\usepackage{xcolor}
\usepackage[normalem]{ulem}
\useunder{\uline}{\ul}{}
\usepackage{bm}
\usepackage{paralist}

\newcommand{\Q}{{\operatorname{quant}}}
\newcommand{\method}{\mbox{\textsc{InvarExplore}}}

\author{%
  Yuqiao Wen$^{1,}$\thanks{Project done during Mitacs internship at RBC Borealis.} \quad Yanshuai Cao$^2$ \quad Lili Mou$^{1,3}$\\
  $^1$Dept. Computing Science \& Alberta Machine Intelligence Institute (Amii), University of Alberta \\
  $^2$RBC Borealis \quad 
  $^3$Canada CIFAR AI Chair \\
 \texttt{yq.when@gmail.com}\\\texttt{yanshuai.cao@borealisai.com}\quad\texttt{doublepower.mou@gmail.com}
}

\newcommand{\mytitle}{
Exploring Model Invariance with Discrete Search \\ for Ultra-Low-Bit Quantization}
\title{\mytitle}

\usepackage{amsmath,amsfonts,bm,xspace,multirow}
\definecolor{myred}{RGB}{255,127,127}
\definecolor{mygreen}{RGB}{127,255,127}

\def\1{\bm{1}}

\DeclareMathAlphabet{\mathsfit}{\encodingdefault}{\sfdefault}{m}{sl}
\SetMathAlphabet{\mathsfit}{bold}{\encodingdefault}{\sfdefault}{bx}{n}

\DeclareMathOperator{\diag}{diag}

\theoremstyle{plain}

\begin{document}

\maketitle

\begin{abstract}
Large language models have been increasing in size due to their success in a wide range of applications.
This calls for a pressing need to reduce memory usage to make them more accessible.
Post-training quantization is a popular technique which uses fewer bits (e.g., 4--8 bits) to represent the model without retraining it. However, it remains a challenging task to perform quantization in an ultra-low-bit setup (e.g., 2 bits).
In this paper, we propose \textsc{InvarExplore}, a unified framework that systematically explores different model invariance at the same time, allowing us to take advantage of the synergy between each type of invariance.
Importantly, \method\ features a discrete search algorithm that enables us to explore permutation invariance, which is under-studied as it cannot be optimized with gradient-based methods.
Results show that \textsc{InvarExplore} is compatible with existing state-of-the-art methods, achieving an add-on performance improvement over strong competing methods.
\end{abstract}

\section{Introduction}\label{sec:intro}

Post-training quantization aims to reduce the number of bits used by a large neural model by lowering the precision of the weights after the model has been trained~\cite{han2016deepcompression}.
It has been gaining growing interest, as pretrained large language models (LLMs) such as ChatGPT~\citep{chatgpt} and Llama~\citep{dubey2024llama} are getting increasingly larger. 

While quantization reduces the model size, it also causes rounding errors and typically leads to performance degradation.
This is especially challenging in the ultra-low-bit setting, for example, using only two or three bits to represent a model weight. 
A promising research direction for quantization is to find an invariant model of the pretrained LLM that reduces rounding errors after quantization and preserves performance.

Existing work finds an invariant model by either manually designed heuristics or gradient-based methods.
\citet{awq} reduce rounding errors by weight scaling and handcraft how much to scale based on activation values.
\citet{omniquant} propose block-wise error minimization to find a scale-invariant model through gradient updates.
\citet{spinquant} also use gradient-based methods and find orthogonal transformations matrices by directly optimizing the cross-entropy loss.
One caveat for computing the gradient is that the quantization function has zero gradient almost everywhere, so they approximate it with straight-through estimation~\citep{bengio2013estimating}. 
In addition, the optimization process is often constrained, and ad hoc treatments are required to ensure that the gradient-updated transformation satisfies the desired form. For example, \citet{spinquant} apply the Cayley SGD method to keep the weight matrix orthogonal.

In this paper, we propose a general framework, called \method, which can explore different types of invariant transformations in neural networks to improve quantization performance.
Specifically, we observe that an invariant model can be obtained by applying a transformation and its inverse to neighboring linear blocks in the network, and the previous scaling method~\citep{awq} is a special case under our framework.
\method\ further allows us to explore permutation and rotation invariance, which are not typically studied in previous literature, especially since permutation is non-differentiable and cannot be trained using gradient-based methods. It is also interesting to notice that, although permutation is a special case of orthogonal transformations, previous gradient-based optimization~\citep{spinquant} cannot easily explore permutation invariance, as permutation creates symmetric local optima in a non-convex fashion. In addition, different types of invariance can be jointly explored under our \method\ framework, which further improves quantization performance.

To address the difficulty of non-differentiability, we propose an activation-guided discrete search algorithm to optimize the invariant transformation under our framework.
At each step of our search algorithm, we sample a combination of permutation, rotation, and scaling for a given layer of the LLM; the updated model is then evaluated based on how well it matches the activation of the original model as well as its perplexity on a fixed, small calibration set (e.g., containing only 512 tokens, which is shorter than the context window).
We adopt hill climbing and only accept changes that improve the model.
Notably, \textsc{InvarExplore} is a lightweight quantization method, as it only requires forward passes for a small number of tokens on a quantized model.

We conducted experiments on two language modeling tasks and six natural language understanding tasks.
Results show that \textsc{InvarExplore} is orthogonal to existing quantization methods, allowing us to build on top of them to achieve an add-on performance improvement.

In summary, the main contributions of this paper are threefold: 1) we propose \method, a general framework that explores different types of invariant transformations; 2) we design an activation-guided discrete search algorithm, which is specifically suitable for the under-explored permutation invariance; and 3) we conduct extensive experiments to evaluate our approach.

\section{Related Work}

\textbf{Integer quantization.} In this paper, we focus on integer quantization, where model weights are stored as integers~\citep{Jacob_2018_CVPR,wu2020integer}.
This is one of the most popular quantization schemes because it offers strong performance, high memory savings, as well as hardware efficiency with integer-only arithmetic during inference in certain scenarios~\citep{Jacob_2018_CVPR,awq}.
Earlier work in this regime has explored binary networks that can only represent $\pm 1$, or ternary networks which additionally include the $0$ value~\citep{NIPS2015_3e15cc11,lin2015neural}. However, they only work with small models for simple tasks.
Researchers typically use more bits (e.g., 4--8 bits) in integer quantization to tackle real-world tasks~\cite{krishnamoorthi2018quantizing,NEURIPS2022_c3ba4962,gptq}, but the ultra-low-bit setup remains challenging. 

Low-bit integers are usually unable to cover the range of real-valued weights. Therefore, a scale parameter is introduced to map the low-bit integer range to the original weight range~\cite{zhou2016dorefa,Jacob_2018_CVPR}. Also, it is important to keep the exact 0 value for weights, so a zero-point parameter is also introduced~\cite{Jacob_2018_CVPR}. Such offset and scaling are applied at the granularity of groups (contiguous weights in a matrix). Although the scale and zero-point parameters use additional memory, they drastically improve performance and are commonly used in modern integer quantization studies~\citep{gptq,awq,omniquant}.

One direction of integer quantization is to quantize weights sequentially~\cite{obq,gptq}, inspired by the classic pruning methods~\citep{obs}. The idea is that, given already quantized weights, the rest weights are finetuned to compensate the quantization error before also being quantized. However, the weight updates are computed in a closed form based on second-order gradient information; this is done for each layer separately, which does not consider the dependencies among layers.

An alternative technique in integer quantization is to leverage model invariance, which adjusts the parameters for better quantization performance without impacting the (un-quantized) model's behavior. Previous work has explored various heuristics. For example, \citet{awq} determine the scaling coefficients based on the magnitude of the activations. \citet{lin2024duquant} manually design a zigzag permutation pattern in an attempt to distribute outlier weights throughout the network.
On the other hand, researchers also use gradient-based methods to learn scaling coefficients~\citep{omniquant} and orthogonal transformations~\citep{spinquant}.
However, this requires a straight-through gradient estimation and special treatment for the constraint, as mentioned in Section~\ref{sec:intro}.

Different from previous work, our paper proposes the  \method\ framework that explores permutation, scaling, and rotation invariance for integer quantization. We further propose a discrete search algorithm that can optimize different types of invariance jointly. 

\textbf{Other representations for quantization.} There are also other ways to represent model weights in a low-bit fashion, such as floating-point quantization~\citep{10.1145/103162.103163,micikevicius2018mixed} and vector quantization~\citep{1162229,gong2015compressing}.

Floating-point quantization mostly follows the widely adopted IEEE 754 standard~(\citeyear{4610935}), where values are represented with a sign bit for polarity, exponent bits for magnitude, and mantissa bits for precision.
The exponent bits enable the floating-point representation to cover a wide range of values, which is especially valuable during training~\citep{NEURIPS2018_335d3d1c}.
To save memory, \citet{micikevicius2018mixed} propose mixed-precision training, which has inspired a line of research focusing on designing different bit representations suitable for training~\citep{NEURIPS2018_335d3d1c,NEURIPS2019_65fc9fb4,NEURIPS2020_13b91943}.
Researchers have also explored adaptive methods, which dynamically adjust the floating-point representation~\citep{NIPS2017_a0160709,Liu_2021_ICCV}. However, these methods usually lack hardware support.
In general, floating point is not suitable for the ultra-low-bit setting of our paper, as we cannot afford to allocate any bits for the exponent.

On the other hand, vector quantization compresses model weights by storing a set of representative vectors, also known as a codebook~\citep{gong2015compressing,han2016deepcompression}.
Recent work such as QuiP\#~\citep{quipsharp} and AQLM~\citep{aqlm} show promising results in this direction, but this is beyond the scope of this paper.

\textbf{Weight-only and weight--activation quantization.}
Quantization can be applied to weights only or both weights and activations.
As shown by previous work, quantizing activation is even harder than quantizing weights, because of the emergence of outlier features in large language models~\citep{NEURIPS2022_c3ba4962,smoothquant}. Weight--activation quantization may use hardware integer arithmetic for fast inference, but it is significantly more challenging, with state-of-the-art methods still requiring 4--6 bits~\citep{wei-etal-2023-outlier,yuan2023rptq,omniquant}. Notice that, although our paper focuses on the weight-only setup for ultra-low-bit quantization, it does not use more memory when storing the model because the activation values are temporary variables during inference.

\section{Approach} \label{sec:approach}

We provide the background for quantization in Subsection~\ref{appr:quantization}. Our \textsc{InvarExplore} framework and the activation-guided discrete search algorithm are presented in Subsection~\ref{appr:invariances}.

\subsection{Quantization Background} \label{appr:quantization}

We adopt the standard asymmetric integer group quantization~\citep{Jacob_2018_CVPR}, where weight matrices are divided into groups of contiguous memory space.
In particular, each group is quantized separately based on a scale parameter that specifies the step size between integer values. Asymmetric quantization allows a zero-point parameter to specify the (integer) offset for the zero value.

Let $\mathbf W_g \in \mathbb R^G$ be the $g$th group containing $G$ contiguous parameters of the weight matrix. Asymmetric integer quantization has the following form
\begin{align}
    \Q(\mathbf W_g) &= \operatorname{round}(\mathbf W_g / s_g) + z_g
\end{align}
where $\Q$ is the quantization function, $s_g$ is the FP16 scale parameter, and $z_g$ is an integer parameter specifying the zero point.

Intuitively, we would like to map the maximum and minimum (un-quantized) values with the largest and smallest integers.
This leads to the closed-form solutions for the scale and zero-point parameters
\begin{align}
    s_g &= \frac{\max(\mathbf W_g) - \min(\mathbf W_g)}{q_{\text{max}} - q_{\text{min}}} \label{eqn:scale} \\
    z_g &= \operatorname{round}(q_\text{min} - \min( {\mathbf W_g}) / s_g) \label{eqn:zpoint}
\end{align}
where $q_{\text{max}}$ and $q_{\text{min}}$ are the largest and smallest numbers that can be represented by the integer type (depending on the number of bits).

To dequantize the integer representations, we can simply subtract the zero point and multiply it by the scale
\begin{align}
    \operatorname{dequant}(\Q(\mathbf W_g)) = s_g \cdot (\Q(\mathbf W_g) - z_g)
\end{align}

The key challenge for quantization, as we can see from Eqns.~\eqref{eqn:scale} and \eqref{eqn:zpoint}, is that the quantization parameters are very sensitive to outliers,  because they heavily depend on the maximum and minimum weights of a group.
These outlier weights cause the scale parameter $s_g$ to be large, which in turn increases the expected quantization error.
Our goal is to use invariant transformations to alleviate this and ultimately improve quantization performance.

\subsection{Our Proposed \textsc{InvarExplore}} \label{appr:invariances}

In this paper, we propose \textsc{InvarExplore}, a framework that explores different types of invariant transformations in neural networks to improve quantization performance. 
The key intuition is that invariant transformations do not alter a model's output if it is not quantized, but they may have different quantization performance because $\operatorname{round}(\cdot)$ is not invertible in the $\Q$ and $\operatorname{dequant}$ functions. Compared with previous work~\citep{awq,smoothquant,ashkboos2024quarot}, our \method\ framework explores permutation and rotation invariance, which can be combined with the previously used scaling invariance.

Specifically, given a model $\mathcal M$ and its trained parameters $\bm\theta_0$, our objective can be formulated as a constrained optimization problem
\begin{align}
     \operatorname*{minimize}_{\bm\theta} & \quad \mathcal L(\mathbf X, \Q(\bm\theta)) \label{eq:loss}\\
     \text{subject to} &\quad \mathcal M(\mathbf x; \bm\theta) = \mathcal M(\mathbf x; \bm\theta_0),\; \forall \mathbf x \label{eqn:constraint}
\end{align}
where $\mathcal L$ is a loss function such as the cross-entropy loss, and $\mathbf X$ is a small batch of calibration data.
Here, Eqn.~\eqref{eqn:constraint} is the invariance constraint that holds for every possible input $\mathbf x$, which includes all samples (not just the ones in the batch). This ensures our search space only contains parameters that yield the same model as the original one.

Unfortunately, Eqn.~\eqref{eqn:constraint} cannot be easily satisfied, because the equality has to hold for all inputs.
Therefore, we restrict our search space to invariant transformations in the feed-forward blocks of the Transformer architecture:
\begin{align}
    \mathbf z = \mathbf W_{\text{down}}f(\mathbf W_\text{up} \mathbf x + \mathbf b_{\text{up}}) + \mathbf b_{\text{down}}
    \label{eqn:ff}
\end{align}
where the $\mathbf W$s and $\mathbf b$s are the weights and biases of the up and down projections, and $f$ is the activation function such as ReLU~\citep{agarap2018deep}.

We consider the following transformations.

\textbf{Permutation.} If we change the order of the neurons in a layer, the network remains invariant. Let $\mathbf P$ be a permutation matrix, i.e., every element is binary and every row/column has exactly one element that is $1$.  The transpose $\mathbf P^\top$ cancels out the permutation effect, as $\mathbf P^\top\mathbf P=\mathbf I$ must hold.

Multiplying $\mathbf P$ to $\mathbf W_\text{up}$ and $\mathbf b_\text{up}$, and $\mathbf P^\top$ to $\mathbf W_\text{down}$ on the appropriate side, we have 
\begin{align}
    & (\mathbf W_{\text{down}}\mathbf P^\top) f((\mathbf P \mathbf W_\text{up}) \mathbf x + \mathbf P \mathbf b_{\text{up}}) + \mathbf b_{\text{down}} \\
    = &\mathbf W_{\text{down}} (\mathbf P^\top \mathbf P) f(\mathbf W_\text{up} \mathbf x + \mathbf b_{\text{up}}) + \mathbf b_{\text{down}} = \mathbf z \label{eqn:pdistr} 
\end{align}
which is the same as the original model \eqref{eqn:ff}.
Here, Eqn.~\eqref{eqn:pdistr} follows because permutation can be equivalently applied before or after the activation function, i.e., $f(\mathbf x \mathbf P) = f(\mathbf x) \mathbf P$.

This gives us the permuted parameters
\begin{align}
    \overline{\mathbf W}_\text{up}  &= \mathbf P \mathbf W_\text{up}, \quad \;\overline{\mathbf b}_\text{up} = \mathbf P \mathbf b_\text{up} \\
     \overline{\mathbf W}_\text{down} &= \mathbf W_\text{down} \mathbf P^\top
\end{align}
Note that the transformed parameters can be efficiently obtained by indexing the original parameters, rather than matrix--matrix multiplications. 

Permutations per se are non-differentiable, but they are special cases of orthogonal transformations, which are differentiable and have been studied  in previous work~\citep{spinquant}. However, they use gradient-based optimization and it cannot effectively explore permutations, because local minima are formed in a non-convex fashion due to the permutation symmetry.

\textbf{Scaling.} For a certain activation function (such as the ReLU and LeakyReLU), we can obtain an invariant model by scaling the features. Let $\bf s$ be a scaling vector, i.e., $s_i$ is the scaling factor for the $i$th dimension. For linear algebra notations, we may define a scaling matrix $\bf S=\diag(\bf s)$, thus having
\begin{align}
    & (\mathbf W_{\text{down}}\mathbf S^{-1}) f((\mathbf S \mathbf W_\text{up}) \mathbf x + \mathbf S \mathbf b_{\text{up}}) + \mathbf b_{\text{down}} \\
    = &\mathbf W_{\text{down}} (\mathbf S^{-1} \mathbf S) f(\mathbf W_\text{up} \mathbf x + \mathbf b_{\text{up}}) + \mathbf b_{\text{down}} = \mathbf z \label{eqn:pdistr} 
\end{align}

The scaled parameters are given by
\begin{align}
    \overline{\mathbf W}_\text{up}  &= \mathbf S \mathbf W_\text{up}, \quad \;\overline{\mathbf b}_\text{up} = \mathbf S \mathbf b_\text{up} \\
     \overline{\mathbf W}_\text{down} &= \mathbf W_\text{down} \mathbf S^{-1}
\end{align}
In implementation, we do not need to calculate $\bf S$ and $\bf S^{-1}$. Instead, we only need to scale a dimension by $s_i$ and scale it back by its reciprocal $1/s_i$.

Notice that, the scaling invariance only holds for certain functions. Luckily, our experiments use OPT models~\citep{zhang2022opt} with the ReLU function, which satisfies the property. For other activation functions, scaling $\bf W_\text{up}$ and $\bf W_\text{down}$ is not invariant, but can be employed as an approximation. Moreover, scaling invariance can be achieved for $\bf W_\text{down}$ and its subsequent LayerNorm operation, which is beyond the scope of this paper.

\textbf{Rotation.} A rotation can also be represented by a matrix $\bf R$, and its inverse is $\bf R^{\top}$. However, rotations are not invariant for non-linear activation functions, meaning that applying rotation in a similar way does not satisfy Eqn.~\eqref{eqn:constraint}:
\begin{align}
    & (\mathbf W_{\text{down}}\mathbf R^\top) f((\mathbf R \mathbf W_\text{up}) \mathbf x + \mathbf R \mathbf b_{\text{up}}) + \mathbf b_{\text{down}} \label{eqn:rtransform}  \\
    \ne &\mathbf W_{\text{down}} (\mathbf R^\top \mathbf R) f(\mathbf W_\text{up} \mathbf x + \mathbf b_{\text{up}}) + \mathbf b_{\text{down}} = \mathbf z 
\end{align}
Nevertheless, if the rotation degree is small, then the values can be mostly recovered by an inverse rotation.
We empirically verify this in our pilot study: the original 13B OPT model achieves a cross-entropy loss of $2.31528$ on the WikiText-2 dataset; with our rotations, the cross-entropy is $2.31525$, which demonstrates a difference of only $0.001\%$. 
Therefore, we can think of Eqn.~\eqref{eqn:rtransform} as an approximate invariant model, leading to our rotated parameters
\begin{align}
    \overline{\mathbf W}_\text{up}  &= \mathbf R \mathbf W_\text{up}, \quad \;\overline{\mathbf b}_\text{up} = \mathbf R \mathbf b_\text{up} \\
     \overline{\mathbf W}_\text{down} &= \mathbf W_\text{down} \mathbf R^\top
\end{align}

To construct the rotation, we partition a $d$-dimensional space (assuming $d$ is even) into pairs of two dimensions; this simplified treatment is similar to that for rotary embeddings~\citep{roformer}. Formally, $\bf R$ can be represented by a block diagonal matrix given rotation angles $\phi_1, \cdots, \phi_{d/2}$
\begin{align}
    \mathbf R = 
    \scalebox{0.8}{$
    \begin{bmatrix}
    \cos \phi_1 & -\sin \phi_1 & \cdots & 0 & 0 \\
    \sin \phi_1 & \cos \phi_1 & \cdots & 0 & 0 \\
    \vdots & \vdots & \ddots & \vdots & \vdots \\
    0 & 0 & \cdots & \cos \phi_{d/2} & -\sin \phi_{d/2} \\
    0 & 0 & \cdots & \sin \phi_{d/2} & \cos \phi_{d/2}
    \end{bmatrix}
    $}
\end{align}
Our rotation transformation is different from~\citet{spinquant}, who perform orthogonal transformation but call it rotation. Rigorously speaking, orthogonal transformations need not be rotations.

Overall, the proposed \method\ allows us to explore permutation, scaling, and rotation, which can be further combined as 
\begin{align}
    \overline{\mathbf W}_\text{up}  &= \mathbf P \mathbf S \mathbf R \mathbf W_\text{up}, \quad \;\overline{\mathbf b}_\text{up} = \mathbf P \mathbf S \mathbf R \mathbf b_\text{up} \\
     \overline{\mathbf W}_\text{down} &= \mathbf W_\text{down}  \mathbf R^\top \mathbf S^{-1} \mathbf P^\top
\end{align}
Here, the order of the transformations can be interchanged, as long as they are properly inverted in the down projection.
In our implementation, we do not store $\mathbf P$, $\mathbf S$, and $\mathbf R$ as matrices; rather, we store them as a permutation vector $\bm \pi$, a scale coefficient vector $\mathbf s$, and a rotation degree vector $\bm \phi$.

\textbf{Activation-guided discrete search.} With the proposed transformation, our optimization problem~\eqref{eq:loss} is reduced to the search of permutation, scaling, and rotation matrices. 
Despite this, it still cannot be optimized by gradient-based methods because the transformation (such as permutation) may impose binary constraints. Moreover, the $\Q$ function produces zero gradient almost everywhere, making the optimization difficult in general.

To this end, we propose to optimize the above-mentioned transformations through discrete search, using the cross-entropy loss and an activation-matching loss as our objective.
Let $\mathbf H$ and $\mathbf H_0$ be the activations of the quantized model and the original model, respectively. Our loss is
\begin{align}
    \mathcal L(\mathbf X, \Q(\bm\theta)) = \operatorname{CE}(\mathbf X, \Q(\bm\theta)) + \alpha \operatorname{MSE}(\mathbf H, \mathbf H_0) \label{eqn:obj}
\end{align}
where $\operatorname{CE}$ refers to the standard cross-entropy loss; $\operatorname{MSE}$ computes the mean squared error, guiding the search process based on the activation values of hidden layers; and $\alpha$ is a hyperparameter balancing the two terms. 

To optimize the loss, we adopt a hill-climbing algorithm.  
At each step, we sample a transformation that combines permutation, rotation, and scaling.
For permutation, we simply reshuffle the neurons in a layer; for scaling and rotation, we adopt random walk and sample the scale and rotation degree based on a Gaussian centered at the current values.
Then, we quantize the model after applying the transformation. If the overall loss is lower, we accept the sampled transformation; otherwise, we reject it.
The process is detailed in Algorithm~\ref{alg:search}.

In practice, we only permute, scale, and rotate a subset of the weights in a layer for each proposed update, and the size of the subset acts as a step size that controls how much we move in the parameter space. A larger step size will result in a lower acceptance rate, while a smaller one will lead to less change in the parameters. We find changing $10\%$ of the neurons within a layer at a time generally works well, and we adopt this for all model sizes.

\begin{center}
\begin{minipage}{0.6\textwidth}

\begin{algorithm}[H]
    \caption{Activation-Guided Discrete Search}
    \label{alg:search}
    \KwInput{
    $\mathbf X$: calibration data; \\
    $\bm\theta_0$: initial model parameters \\
    $\sigma_\text{r}$, $\sigma_\text{s}$: standard deviation for scaling and rotation \\
    $\alpha$: a balancing hyperparameter \\
    }
    $\mathbf H \gets \mathcal M(\mathbf X, \bm \theta_0)$ \\ 
    $\mathbf H_0 \gets \mathcal M(\mathbf X, \Q (\bm \theta_0))$ \\ 
    $\mathcal L_\text{best} \gets \mathcal L_{\text{KL}}(\mathbf X, \Q(\bm\theta_0)) + \alpha \operatorname{MSE}(\mathbf H, \mathbf H_0)$ \\
    $\bm \theta \gets \bm \theta_0$ \\
    $\triangleright$ Initialization for each layer $l$ \\
    \For{$l = 1, \cdots, L$}{
        {$\bm \pi_l \gets [0, 1, 2, \cdots] \in \mathbb R^{d}$ \Comment{No permutation}} \\
        $\mathbf s_l \gets [1, 1, \cdots] \in \mathbb R^{d}$ \Comment{No scaling} \\
        $\bm \phi_l \gets [0, 0, \cdots] \in \mathbb R^{d/2}$ \Comment{No rotation} \\
    }
    \For{$t = 1, 2, \cdots$}{
        Sample a layer $l$ \\
        $\bm \pi' \gets \operatorname{shuffle}(\bm \pi_l)$ \\
        $\mathbf s' \sim \mathcal N(\mathbf s_l, \sigma_{\text{s}}^2)$ \\
        $\bm \phi' \sim \mathcal N(\bm \phi_l, \sigma_{\text{r}}^2)$ \\
        $\bm\theta' \gets \operatorname{apply\_transformation}(\bm\theta, \bm \pi', \mathbf s', \bm \phi', l)$ \\
        $\mathcal L' \gets \mathcal L_{\text{KL}}(\mathbf X, \Q(\bm\theta')) + \alpha \operatorname{MSE}(\mathbf H, \mathbf H_0)$ \\
        \If{$\mathcal L' < \mathcal L_{\textnormal{best}}$}{
            $\bm \theta \gets \bm \theta', \mathcal L_{\text{best}} \gets \mathcal L'$ \\
            $\bm \pi_l \gets \bm \pi', \mathbf s_l \gets \mathbf s', \bm \phi_l \gets \bm \phi'$
        }
    }
\end{algorithm}

\end{minipage}
\end{center}

\section{Experiments}

We provide the settings of our experiments in Subsection~\ref{subsec:settings}, and present our results in Subsection~\ref{subsec:results}.

\begin{table*}[]
\centering
\resizebox{\textwidth}{!}{
\begin{tabular}{|l||cccc|cccc||cccc|}
\hline
Dataset & \multicolumn{4}{c|}{WikiText-2} & \multicolumn{4}{c||}{Colossal Clean Crawled Corpus (C4)} & \multicolumn{4}{c|}{Reasoning Tasks} \\ \hline
Model size & 1.3B & 2.7B & 6.7B & 13B & 1.3B & 2.7B & 6.7B & 13B & 1.3B & 2.7B & 6.7B & 13B \\ \hline\hline
FP16 & 14.62 & 12.47 & 10.86 & 10.13 & 15.73 & 14.10 & 12.54 & 11.93 & 55.50 & 59.71 & 64.62 & 66.11 \\
RTN & 1.30e4 & 5.68e4 & 7.82e3 & 7.65e4 & 7.87e3 & 3.91e4 & 5.50e3 & 3.01e4 & 36.22 & 39.86 & 35.51 & 36.41 \\ \hline
GPTQ~\citep{gptq} & 396.28 & 206.06 & 30.31 & 40.20 & 186.67 & 111.27 & 27.80 & 51.85 & 40.93 & 44.39 & 50.18 & 48.57 \\
+\textsc{InvarExplore} & \textbf{213.33} & \textbf{103.09} & \textbf{27.83} & \textbf{29.49} & \textbf{117.66} & \textbf{67.04} & \textbf{26.05} & \textbf{26.66} & \textbf{42.94} & \textbf{45.90} & \textbf{50.96} & \textbf{51.95} \\ \hline
AWQ~\citep{awq} & 53.57 & 80.24 & 26.11 & 35.89 & 46.82 & 87.14 & 26.33 & 37.36 & 41.69 & 46.45 & 52.91 & 51.85 \\
+\textsc{InvarExplore} & \textbf{41.89} & \textbf{50.07} & \textbf{22.65} & \textbf{26.26} & \textbf{38.04} & \textbf{40.05} & \textbf{23.62} & \textbf{27.00} & \textbf{44.33} & \textbf{49.56} & \textbf{53.95} & \textbf{55.13} \\ \hline
OmniQuant~\citep{omniquant} & 23.96 & 18.15 & 14.43 & 12.94 & 29.24 & 22.55 & 17.86 & 16.06 & 47.12 & \textbf{50.62} & 56.74 & 59.55 \\
+\textsc{InvarExplore} & \textbf{23.33} & \textbf{17.94} & \textbf{14.20} & \textbf{12.73} & \textbf{28.54} & \textbf{22.36} & \textbf{17.39} & \textbf{15.78} & \textbf{47.58} & \textbf{50.62} & \textbf{56.87} & \textbf{59.67} \\ \hline
\end{tabular}
}
\caption{Main results. For language modeling on WikiText-2 and C4 corpora, we report the perplexity score (the lower the better). For reasoning, we report the accuracy averaged across six tasks (the higher the better). Detailed results are shown in Appendix~\ref{app:reasoning-full}.}
\label{tab:main}
\end{table*}

\subsection{Settings} \label{subsec:settings}

\textbf{Evaluation datasets.} 
We evaluate our \method\ on both language modeling and a selection of popular reasoning tasks.
For language modeling, we compare the perplexity scores on the widely used WikiText-2 dataset~\citep{wikitext} and the Colossal Clean Crawled Corpus~\citep[C4,][]{c4}.
For reasoning, we adopt the popular few-shot evaluation harness.\footnote{\url{https://github.com/EleutherAI/lm-evaluation-harness}}
Specifically, we follow \citet{omniquant} and test on the following tasks

\quad$\bullet$ \underline{ARC}~\citep{clark2018think}: a benchmark consisted of multiple choice questions on grade school science.
The authors partition their data into an easy subset and a challenging subset, denoted by ARC-E and ARC-C, respectively. \\
\mbox{\quad}$\bullet$ \underline{BoolQ}~\citep{clark-etal-2019-boolq}: a reading comprehension dataset focusing on yes/no questions. \\
\mbox{\quad}$\bullet$ \underline{HellaSwag}~\citep{zellers-etal-2019-hellaswag}: a natural language inference benchmark for common sense reasoning. \\
\mbox{\quad}$\bullet$ \underline{PIQA}~\citep{bisk2020piqa}: a common sense reasoning benchmark focusing on interactions with the physical world. \\
\mbox{\quad}$\bullet$ \underline{WinoGrande}~\citep{sakaguchi2020winogrande}: a pronoun resolution dataset, where the task is to choose the correct entity that a pronoun refers to given two options.

\textbf{Experimental details.} We adopt the Open Pretrained Transformer (OPT) family of LLMs, ranging from 1.3B to 13B parameters~\citep{zhang2022opt}, which have been used in many previous studies on quantization~\citep{gptq,awq,omniquant}.
Our LLM is prompted by five-shot in-context learning samples, similar to the setting in~\citet{pmlr-v235-li24bb}. In-context samples are needed, as we observe low performance in some cases with zero-shot prompting due to the ultra-low-bit nature of our study.

We adopt the 2-bit setting with a group size of 128 for our main experiment, which is typically included in previous studies but not as their main focus~\citep{awq,omniquant}.  Our analysis also experimented with the 3-bit setting.

In our algorithm, we need a small calibration dataset to optimize the objective in Eqn.~\eqref{eq:loss}. In particular, we use 32 sequences from the Pile corpus~\citep{gao2020pile} in our main experiments, each sequence containing 512 tokens. Our calibration set is similar to \citet{awq} but smaller than \citet{gptq} and \citet{omniquant}. Notably, a sequence with 512 tokens is much shorter than the context window of modern LLMs (e.g., OPT has a context of 2048 tokens), which suggests that our calibration with the short sequences is memory- and computation-efficient due to the quadratic scaling of attention in Transformers. In our analysis, we will show that our approach can also work with only one sequence.

In our algorithm, the balancing hyperparameter $\alpha$ in Eqn.~\eqref{eqn:obj} is chosen such that the cross-entropy loss is ten times more important than the activation mean squared error at the beginning of training. For scaling and rotation, we need the random-walk standard deviations $\sigma_s$ and $\sigma_r$ to sample new candidate values (Algorithm~\ref{alg:search}); they are set to 1e-2 and 1e-5, respectively. The rotation standard deviation $\sigma_r$ is much smaller because we observe high variance if the rotation degrees are large.
The above hyperparameters are obtained via a grid search during our pilot study based on the loss on the calibration set. We use the same hyperparameters across all settings and datasets.

\subsection{Results and Analyses} \label{subsec:results}

\begin{table*}[t]
\centering
\resizebox{\textwidth}{!}{
\begin{tabular}{|l|cc|ccccccc|}
\hline
Metric & \multicolumn{2}{c|}{Perplexity$^\downarrow$} & \multicolumn{7}{c|}{Accuracy$^\uparrow$} \\ \hline
Dataset & WikiText-2 & C4 & ARC-C & ARC-E & BoolQ & HellaS & PIQA & WinoG & Avg \\ \hline\hline
FP16 & 10.13 & 11.93 & 40.10 & 71.51 & 68.47 & 70.59 & 77.26 & 68.75 & 66.11 \\ \hline
RTN & 76479.03 & 30125.99 & 25.60 & 26.47 & 38.23 & 25.59 & 50.33 & 52.25 & 36.41 \\
AWQ & 35.89 & 37.36 & 26.37 & 49.75 & 62.81 & 48.65 & 68.44 & 55.09 & 51.85 \\
+\textsc{InvarExplore}-Permuation & 29.95 & {\ul 31.34} & \textbf{28.50} & 53.49 & {\ul 62.78} & {\ul 52.12} & {\ul 69.15} & {\ul 56.59} & 53.77 \\
+\textsc{InvarExplore}-Scaling & 33.19 & 36.05 & 28.07 & 51.18 & 62.87 & 50.75 & 68.61 & 56.04 & 52.92 \\
+\textsc{InvarExplore}-Rotation & {\ul 30.21} & 30.94 & \textbf{27.90} & {\ul 53.11} & 63.24 & 52.84 & 70.40 & 58.17 & {\ul 54.28} \\
+\textsc{InvarExplore} (All) & \textbf{26.26} & \textbf{27.00} & {\ul 28.16} & \textbf{54.63} & \textbf{63.94} & \textbf{54.51} & \textbf{70.95} & \textbf{58.56} & \textbf{55.13} \\ \hline
\end{tabular}
}
\caption{Ablation study for permutation, scaling, and rotation with the 13B OPT model.}
\label{tab:ablation}
\end{table*}

\begin{table*}[]
\centering
\resizebox{\textwidth}{!}{
\begin{tabular}{|c|c|c|l||cc||ccccccc|}
\hline
Bit & Group & \multicolumn{1}{l|}{Bits/Param} & \multicolumn{1}{c||}{Method} & WikiText-2 & C4 & ARC-C & ARC-E & BoolQ & HellaS & PIQA & WinoG & Avg \\ \hline\hline
- & - & 16 & FP16 & 10.13 & 11.93 & 40.10 & 71.51 & 68.47 & 70.59 & 77.26 & 68.75 & 66.11 \\ \hline
\multirow{2}{*}{1} & \multirow{2}{*}{64} & \multirow{2}{*}{1.25} & AWQ & 93672.90 & 43563.64 & 25.94 & 26.14 & \textbf{37.83} & 26.02 & 49.89 & 49.72 & 35.92 \\
 &  &  & +\method & \textbf{6823.46} & \textbf{4561.87} & \textbf{26.28} & \textbf{26.64} & \textbf{37.83} & \textbf{26.27} & \textbf{50.05} & \textbf{51.22} & \textbf{36.38} \\ \hline
\multirow{2}{*}{2} & \multirow{2}{*}{64} & \multirow{2}{*}{2.25} & AWQ & 20.26 & 23.54 & 29.44 & 54.59 & \textbf{64.43} & 52.72 & 69.42 & 55.64 & 54.37 \\
 &  &  & +\method & \textbf{19.02} & \textbf{20.96} & \textbf{32.00} & \textbf{59.22} & 64.31 & \textbf{56.40} & \textbf{71.27} & \textbf{59.04} & \textbf{57.04} \\ \hline
\multirow{2}{*}{2} & \multirow{2}{*}{128} & \multirow{2}{*}{2.125} & AWQ & 35.89 & 37.36 & 26.37 & 49.75 & \textbf{62.81} & 48.65 & 68.44 & 55.09 & 51.85 \\
 &  &  & +\method & \textbf{26.26} & \textbf{27.00} & \textbf{28.16} & \textbf{54.63} & \textbf{63.94} & \textbf{54.51} & \textbf{70.95} & \textbf{58.56} & \textbf{55.13} \\ \hline
\multirow{2}{*}{3} & \multirow{2}{*}{128} & \multirow{2}{*}{3.125} & AWQ & 10.81 & \textbf{12.76} & \textbf{38.99} & \textbf{69.53} & 69.42 & 68.43 & \textbf{76.44} & \textbf{67.25} & \textbf{65.01} \\
 &  &  & +\method & \textbf{10.77} & 12.77 & 38.74 & 69.49 & \textbf{69.63} & \textbf{68.73} & \textbf{76.44} & 66.22 & 64.88 \\ \hline
\end{tabular}
}
\caption{Performance of the 13B OPT model with different number of bits and group sizes.}
\label{tab:bits}
\end{table*}

\textbf{Main results.} 
Our \method\ is a general approach and can be applied to existing quantization methods. We consider popular and state-of-the-art methods: \underline{GPTQ} quantizes weights sequentially and adjust the remaining weights to compensate for the quantization error~\citep{gptq}, \underline{AWQ} explores the scaling invariance by heuristics~\citep{awq}, and \underline{OmniQuant} learns the scaling coefficients by gradient updates.
In addition, both AWQ and OmniQuant use weight clipping to alleviate outlier weights.

In our experiments, we replicate GPTQ and AWQ by running their provided codebases, as they do not release checkpoints for the 2-bit setting that we focus on; for OmniQuant, we directly use the publicly available checkpoint. In general, we achieve similar trends as previous work. For example, our replication of OmniQuant matches closely with \citet{omniquant}. This establishes a solid foundation for the rest experiments.

Our main results are presented in Table~\ref{tab:main}, where we show performance on both language modeling and downstream reasoning tasks.
For reasoning, we report the average accuracy across six reasoning tasks; the detailed performance for each task can be found in Appendix~\ref{app:reasoning-full}.

We first examine the standard, un-quantized OPT model with the FP16 representation. 
As seen in Table~\ref{tab:main}, larger models consistently achieve lower perplexity scores and higher accuracies, which is expected. 

Next, we experiment with the round-to-nearest (RTN) baseline that applies the standard integer quantization and rounds the model weights with the nearest integer representation. Our replication results match exactly with those in~\citet{omniquant}. As we see, the simple RTN approach performs poorly for low-bit quantization, as it increases the perplexity by orders of magnitude. 

For modern quantization methods GPTQ, AWQ, and OmniQuant, they significantly outperform the RTN baseline, with OmniQuant achieving the highest performance, followed by the popular AWQ approach, which is in turn followed by the earlier GTPQ approach.

We then apply the proposed \method\ and see that our approach achieves consistent improvement over the state-of-the-art quantization methods. In particular, we achieve substantial improvement over GPTQ and AWQ, for instance, reducing the language modeling perplexity of GPTQ by 30--50\% and increasing the reasoning accuracy by 3 points in the 13B setting.
Our improvement over OmniQuant is more modest, as OmniQuant has already achieved high performance. Nevertheless, our improvement is generally consistent in language modeling and downstream reasoning tasks across model sizes (ranging from 1.3B to 13B).
This shows the effectiveness of \method\ in both language modeling and downstream tasks, highlighting its generality and practical values.

\textbf{Ablation for different transformations.} We analyze the effectiveness of different transformation types, namely, permutation, scaling, and rotation across language modeling and reasoning tasks. Due to the limit of space and computing resources, we consider the AWQ (which our implementation is based on) and the 13B OPT model for our analyses.

As shown in Table~\ref{tab:ablation}, every type of transformation alone outperforms AWQ across all tasks except that the permutation-only variant is 0.03 points lower on the BoolQ dataset (which is understandable due to the large number of tasks and settings). We observe that permutation and rotation yield more improvement than scaling, as AWQ has already applied scaling based on activation values, making further exploration less effective.

Moreover, combining all three transformations improves performance by 3--9 points in language modeling and 1--2 average points on reasoning tasks, compared with individual transformations. This suggests that  \method\ is a general framework that facilitates synergistic exploration of different types of invariance for neural network quantization.

\begin{table*}[]
\centering
\resizebox{\textwidth}{!}{
\begin{tabular}{|l|c|cc|ccccccl|}
\hline
\multicolumn{1}{|c|}{Method} & Extra memory & WikiText-2 & C4 & ARC-C & ARC-E & BoolQ & HellaSwag & PIQA & WinoGrande & Avg \\ \hline
AWQ & - & 35.89 & 37.36 & 26.37 & 49.75 & 62.81 & 48.65 & 68.44 & 55.09 & \multicolumn{1}{c|}{51.85} \\ \hline
+\method\ w/ 0 layers & 0.00 GiB & 29.26 & 30.36 & 27.47 & 53.28 & 63.64 & 53.10 & 70.13 & \textbf{59.43} & 54.51 \\
+\method\ w/ 1 layers & 0.31 GiB & 27.87 & 29.27 & \textbf{29.01} & 54.21 & 64.19 & 54.00 & 70.24 & 56.91 & {\ul 54.76} \\
+\method\ w/ 5 layers & 1.56 GiB & 27.09 & 27.92 & 27.73 & \textbf{54.67} & 62.78 & 54.47 & 69.91 & 58.96 & 54.75 \\
+\method\ w/ 10 layers & 3.13 GiB & \textbf{26.26} & \textbf{27.00} & 28.16 & 54.63 & \textbf{63.94} & \textbf{54.51} & \textbf{70.95} & 58.56 & \textbf{55.13} \\ \hline
\end{tabular}
}
\caption{Analysis on the number of layers used for activation matching for the 13B OPT model. The extra memory needed is based on 32 sequences of 512-token calibration samples.
}
\label{tab:activations}
\end{table*}

\begin{figure*}
    \centering
    \includegraphics[width=0.9\linewidth]{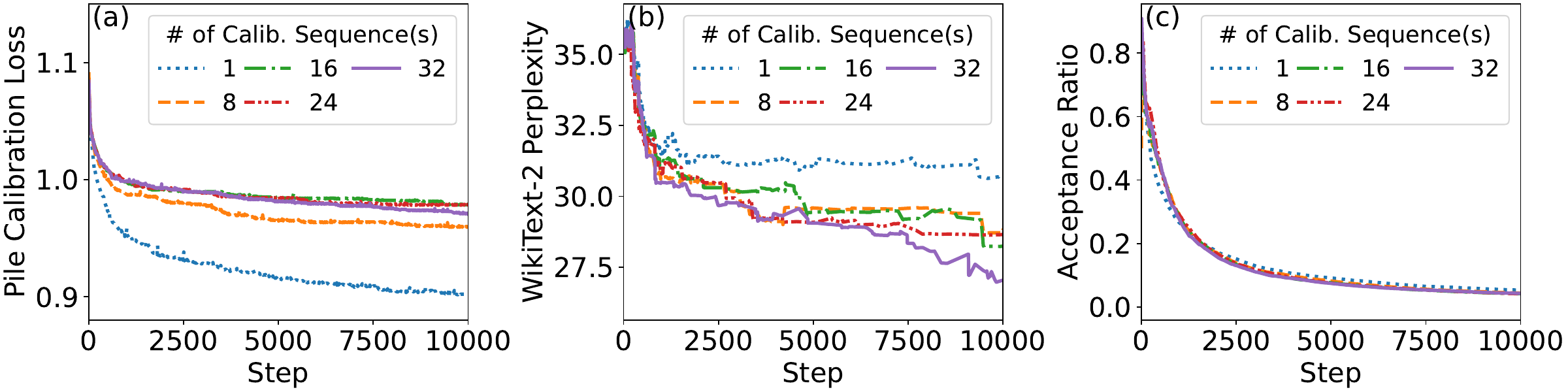}
    \vspace{-5pt}
    \caption{Optimization curves for the 13B OPT model across different numbers of calibration sequences: (a) the calibration loss on the Pile corpus, (b) the perplexity scores on WikiText-2, and (c) the acceptance ratio of sampled transformations.}
    \label{fig:curve}
\end{figure*}

\textbf{Analysis of the number of bits and group size.} We evaluate our \method\ under different quantization settings, including 1-, 2-, and 3-bit quantization using 64 and 128 as the group size.
Results are presented in Table~\ref{tab:bits}.
We omit higher-bit quantization settings, as the performance of existing methods is nearly saturated with the 3-bit setting (only 1 point lower in reasoning than FP16), but the performance of 1-bit and 2-bit settings remains low. This highlights the challenge of the ultra-low-bit setup.

We first compare the effect of the number of bits. Since the 1-bit setting essentially means each weight can only take two values, it becomes extremely challenging and thus we use a smaller group size of 64 (implying a more fine-grained quantization) during the comparison. Still, the 1-bit setting yields very high perplexity and low accuracy for AWQ. Our \method\ significantly reduces the perplexity by an order of magnitude, but it is not enough to recover the reasoning capability. For the 2-bit setting, we see a seemingly small improvement of \url{~}1 point on WikiText-2, but we achieve \url{~}3-point improvement in reasoning. For the 3-bit setting, we see \method\ is similar to AWQ, whose performance is already close to FP16, leaving little room for further improvement.

We also compare 2-bit quantization with different group sizes. As we can see, a smaller group size enables a more fine-grained quantization, yielding a higher performance with a slightly larger memory cost.

Overall, our \method\ works generally well in different quantization settings, achieving higher or similar performance compared with the AWQ approach. In particular, the ultra-low 2-bit setup allows for a memory saving of 85\%, while our approach yields multi-point improvement compared with the baseline, achieving a balance between performance and model size.

\textbf{Effect of activation matching.} Our search objective is shown in Eqn.~\eqref{eqn:obj}, which combines the task-oriented cross-entropy loss and an activation-matching loss. Although the activation matching can be applied to all layers, we find doing so for the 40-layer 13B OPT model requires a large memory, which does not fit in our GPU. Therefore, we restrict the matching for 10 layers in the main experiment and analyze the effect of matched layers in Table~\ref{tab:activations}.

We see that matching more layers generally leads to higher performance, as the activation-matching loss provides guidance throughout the deep neural network. The effect is analogous to intermediate-layer matching in knowledge distillation~\citep{patientkd,jiao-etal-2020-tinybert}. 

Notice that activation matching comes with an overhead of memory during the search process because we need to store the activation values for the calibration samples. When activation matching is disabled (w/ 0 layers), our \method\ still consistently outperforms AWQ across all tasks without any memory overhead. This suggests that an end user---who has enough GPU memory to perform inference with a quantized model---would also have enough memory to apply our method to improve quantization performance.

\textbf{Number of calibration sequences.}
We also study how the size of the calibration dataset affects our approach.
Specifically, we evaluate \method\ using 1, 8, 16, 24, and 32 sequences for calibration, with each sequence containing 512 tokens.
We present the optimization curves for the first 10K steps, which is generally sufficient for \method\ to make significant progress.

As shown in Figures~\ref{fig:curve}a and \ref{fig:curve}b, both the calibration loss and the test perplexity consistently decrease as we increase the search step.
As expected, using fewer calibration sequences generally makes the calibration loss decrease faster, but that also leads to slower improvement on the test set.

We also show the acceptance rate of \method\ in Figure~\ref{fig:curve}c. Recall that our optimization performs a random walk-style hill-climbing search, where a new transformation is accepted if it leads to a lower loss. We see the acceptance rate is around 80\% at the beginning, showing that our search is highly efficient despite using random walk. With more steps, the acceptance rate steadily decreases and eventually flattens out, as the search algorithm empirically converges.

\section{Conclusion}

In this paper, we propose the \textsc{InvarExplore} framework to explore different types of model invariance, including permutation, scaling, and rotation. We also propose a random walk-style hill-climbing search algorithm that works for non-differentiable transformations to improve quantization performance.
Results show that \textsc{InvarExplore} achieves an add-on performance improvement over existing state-of-the-art quantization methods, highlighting the effectiveness and generality of our proposed method.

\textbf{Limitation and future work.} It is noticed in Section~\ref{sec:approach} that our rotation does not provide exact invariance with non-linear activation functions, although the training loss only differs by 0.001\% in our scenario. We observe future opportunities to explore exact rotational invariance in other model architectures, such as low-rank adaptation~(LoRA; \citeauthor{hulora}, \citeyear{hulora}).

\section{Impact Statement}

Our work addresses the ultra-low-bit setup for language model quantization. We believe it will have a positive societal impact, as it makes language models more accessible to ordinary users.

\section*{Acknowledgments}
The research is supported in part by the Natural Sciences and Engineering Research Council of Canada (NSERC), a Mitacs Accelerate project, the Amii Fellow Program, the Canada CIFAR AI Chair Program, an Alberta Innovates Program, and the Digital Research Alliance of Canada (alliancecan.ca).

\bibliography{main}

\vfill
\pagebreak

\appendix

\section{Detailed Results on Reasoning Tasks} \label{app:reasoning-full}

Table~\ref{tab:reasoning-full} presents accuracies for individual reasoning tasks.
As we can see, \method\ is generally effective across different tasks, with different base models of different sizes (58 wins, 11 loses, and 3 ties with two-sided bionimal test yielding a $p$-value of 8e-9).

\begin{table}[t]
\centering
\resizebox{0.8\textwidth}{!}{
\begin{tabular}{|c|l|ccccccc|}
\hline
Model size & \multicolumn{1}{c|}{Method} & Arc-C & Arc-E & BoolQ & HellaG & PIQA & WinoG & Avg \\ \hline
\multirow{8}{*}{1.3B} & FP16 & 29.61 & 59.72 & 59.48 & 54.07 & 71.06 & 59.04 & 55.50 \\
 & RTN & 26.88 & 26.14 & 37.83 & 25.82 & 51.41 & 49.25 & 36.22 \\ \cline{2-9} 
 & GPTQ & \textbf{23.63} & 32.07 & 53.64 & 29.98 & 55.01 & 51.22 & 40.93 \\
 & +\method & 22.87 & \textbf{34.47} & \textbf{57.58} & \textbf{31.04} & \textbf{57.51} & \textbf{54.14} & \textbf{42.94} \\ \cline{2-9} 
 & AWQ & 22.53 & 39.14 & 40.67 & 34.51 & 60.34 & 52.96 & 41.69 \\
 & +\method & \textbf{24.15} & \textbf{40.74} & \textbf{45.47} & \textbf{38.51} & \textbf{63.44} & \textbf{53.67} & \textbf{44.33} \\ \cline{2-9} 
 & OmniQuant & 24.15 & 42.76 & 60.18 & 38.60 & \textbf{63.22} & 53.83 & 47.12 \\
 & +\method & \textbf{24.74} & \textbf{43.69} & \textbf{60.80} & \textbf{39.06} & \textbf{63.22} & \textbf{53.99} & \textbf{47.58} \\ \hline
\multirow{8}{*}{2.7B} & FP16 & 32.94 & 65.11 & 62.66 & 60.69 & 74.92 & 61.96 & 59.71 \\
 & RTN & 24.23 & 25.63 & 62.17 & 25.98 & 51.09 & 50.04 & 39.86 \\ \cline{2-9} 
 & GPTQ & 23.29 & 38.47 & 59.79 & 35.30 & 58.60 & 50.91 & 44.39 \\
 & +\method & \textbf{25.00} & \textbf{39.06} & \textbf{61.68} & \textbf{36.79} & \textbf{59.90} & \textbf{52.96} & \textbf{45.90} \\ \cline{2-9} 
 & AWQ & 23.89 & 42.76 & 60.61 & 33.72 & 64.36 & 53.35 & 46.45 \\
 & +\method & \textbf{24.40} & \textbf{47.85} & \textbf{61.65} & \textbf{42.08} & \textbf{66.21} & \textbf{55.17} & \textbf{49.56} \\ \cline{2-9} 
 & OmniQuant & 26.96 & 52.02 & \textbf{56.91} & 45.25 & \textbf{66.92} & \textbf{55.64} & \textbf{50.62} \\
 & +\method & \textbf{27.30} & \textbf{52.23} & 56.09 & \textbf{45.51} & \textbf{66.92} & \textbf{55.64} & \textbf{50.62} \\ \hline
\multirow{8}{*}{6.7B} & FP16 & 37.03 & 69.70 & 70.18 & 67.95 & 77.04 & 65.82 & 64.62 \\
 & RTN & 25.68 & 25.21 & 37.83 & 26.41 & 49.46 & 48.46 & 35.51 \\ \cline{2-9} 
 & GPTQ & \textbf{27.56} & \textbf{51.52} & 52.94 & 50.38 & 65.45 & 53.20 & 50.18 \\
 & +\method & 27.39 & 50.80 & \textbf{55.90} & \textbf{50.89} & \textbf{66.70} & \textbf{54.06} & \textbf{50.96} \\ \cline{2-9} 
 & AWQ & 28.24 & 55.56 & \textbf{56.42} & 51.06 & 69.37 & 56.83 & 52.91 \\
 & +\method & \textbf{30.20} & \textbf{56.73} & 55.54 & \textbf{52.98} & \textbf{70.29} & \textbf{57.93} & \textbf{53.95} \\ \cline{2-9} 
 & OmniQuant & \textbf{31.83} & 60.44 & \textbf{63.85} & 54.25 & \textbf{71.49} & 58.56 & 56.74 \\
 & +\method & 31.14 & \textbf{60.94} & 63.67 & \textbf{54.60} & 71.44 & \textbf{59.43} & \textbf{56.87} \\ \hline
\multirow{8}{*}{13B} & FP16 & 40.10 & 71.51 & 68.47 & 70.59 & 77.26 & 68.75 & 66.11 \\
 & RTN & 25.60 & 26.47 & 38.23 & 25.59 & 50.33 & 52.25 & 36.41 \\ \cline{2-9} 
 & GPTQ & 27.13 & 46.13 & 57.77 & 41.93 & 63.66 & 54.78 & 48.57 \\
 & +\method & \textbf{28.41} & \textbf{50.00} & \textbf{60.21} & \textbf{50.48} & \textbf{66.70} & \textbf{55.88} & \textbf{51.95} \\ \cline{2-9} 
 & AWQ & 26.37 & 49.75 & 62.81 & 48.65 & 68.44 & 55.09 & 51.85 \\
 & +\method & \textbf{28.16} & \textbf{54.63} & \textbf{63.94} & \textbf{54.51} & \textbf{70.95} & \textbf{58.56} & \textbf{55.13} \\ \cline{2-9} 
 & OmniQuant & 32.68 & 64.73 & \textbf{66.09} & 58.58 & \textbf{73.07} & \textbf{62.12} & 59.55 \\
 & +\method & \textbf{33.28} & \textbf{64.86} & 65.78 & \textbf{59.36} & 72.96 & 61.80 & \textbf{59.67} \\ \hline
\end{tabular}
}
\caption{Performance on six reasoning tasks across OPT models from 1.3B to 13B.}
\label{tab:reasoning-full}
\end{table}

\end{document}